\documentclass[11pt,journal,compsoc,onecolumn]{IEEEtran}
\usepackage{cite}

\ifCLASSINFOpdf
\else
\fi
\ifCLASSOPTIONcompsoc
 \usepackage[font=normalsize,labelfont=sf,textfont=sf]{subfig}
\else
 \usepackage[font=footnotesize]{subfig}
\fi
\ifCLASSOPTIONcompsoc
 \usepackage[font=normalsize,labelfont=sf,textfont=sf]{subfig}
\else
 \usepackage[font=footnotesize]{subfig}
\fi

\usepackage{color}
 \usepackage{amssymb}
\usepackage{amsfonts}
\usepackage{dsfont}
\usepackage{epsfig}

\usepackage{times}
\usepackage{graphicx}
\usepackage{amsmath}
\usepackage[ruled,lined,linesnumbered,boxed]{algorithm2e}

\title{Cascaded Coarse-to-Fine  Deep Kernel Networks for Efficient Satellite Image Change Detection}
\author{Hichem Sahbi \\ $ $ \\ {CNRS, LIP6 Lab, Sorbonne University, Paris}}

\begin{document}

\def \x{{\bf x}}
\def \y{{\bf y}}
\def \k{{\bf \kappa}}
\def \w{{\bf w}}
\def \Y{{\bf Y}}
\def \W{{\bf W}}
\def \G{{\bf G}}
\def \F{{\bf F}}

\maketitle

\begin{abstract}
Deep networks are nowadays becoming popular in many computer vision and pattern recognition tasks. Among these networks, deep kernels are particularly interesting and effective, however, their computational complexity is a major issue especially on cheap hardware resources. \\

In this paper, we address the issue of efficient computation in deep kernel networks. We propose a novel framework that reduces dramatically the complexity of evaluating these deep kernels. Our method is based on a coarse-to-fine cascade of networks designed for efficient computation; early stages of the cascade are cheap and reject many patterns efficiently while deep stages are more expensive and accurate. The design principle of these reduced complexity networks is based on a variant of the cross-entropy criterion that reduces the complexity of the networks in the cascade while preserving all the positive responses of the original kernel network. Experiments conducted -- on the challenging and time demanding change detection task, on very large satellite images -- show that our proposed coarse-to-fine approach is effective and highly efficient.
  
\end{abstract}

\section{Introduction}

\noindent

With the era of big data, there is an exponential growth of image collections in the web and this makes their manual annotation and search completely out of reach. With this growth rate, there is an urgent need for reliable and also efficient automatic solutions able to annotate and search these large collections \cite{cusano2003image,tollari2008comparative,nowak2010reliable,napoleon20102d,carneiro2005formulating,ferecatu2008telecomparistech,makadia2010baselines,boujemaa2004visual,ashley1995automatic,bourdis2011constrained,thiemert2006using,srikanth2005exploiting,thiemert2005applying,sahbi2001accurate}. Visual concept detection  is one of these  major challenges that consists of recognizing and localizing concepts/events into flows of visual contents using variety of machine learning and inference techniques \cite{li2011superpixel,joachims1998text,sahbi2015imageclef,freund1996experiments,sahbi2010context,dietterich2000experimental,sahbi2013cnrs,li2009markov,sahbi2002coarse,jordan1998learning,sahbi2008context,mika1999fisher,sahbi2002face,sahbi2011context,yamato1992recognizing,sahbi2008robust,sahbi2004kernel};  among these techniques, deep and convolutional neural networks are particularly successful (see for instance \cite{krizhevsky2012imagenet,he2016deep,srivastava2015training,szegedy2015going,girshick2014rich,russakovsky2015imagenet,he2015delving,jiang2017exploiting,karpathy2014large,yue2015beyond,deng2014deep,ji20133d,Goodfellowetal2016,jiu2016deep}). Recent breakthroughs and success stories of deep learning -- in vision, pattern recognition and neighboring fields -- are also due to the development  of extremely efficient hardware resources that make running deep learning models on bigdata much more tractable.   However, on widely used cheap hardware devices, deep learning models are still very time demanding and require careful algorithmic design in order to achieve efficient computation while maintaining a high accuracy.\\

\indent Deep learning models usually operate on vectorial data;  the underlying parametric models take vectorial inputs and return discriminatively-trained representations and similarities~\cite{chopra2005learning,taigman2014deepface}. When only relationships (or similarities) between input data are available\footnote{for instance through graphs in social networks.}, deep kernel networks become better alternatives~\cite{strobl2013deep,zhuang2011two,jiu2017nonlinear,jose2013local,wilson2016deep,cho2009kernel,jiu2015semi}.  These networks are defined as recursive multi-layered combinations of standard kernels (e.g., Gaussian, random walks, etc.) that capture simple linear as well as intricate nonlinear relationships between input data. Learning the parameters of  these  networks together with classifiers allows us to achieve deep learning on non-vectorial data\footnote{in the Hilbert space associated to these kernels.} more effectively compared to existing standard kernels as well as shallow multiple kernels~\cite{cristianini2000introduction,bach2004multiple}.    However, the downside of the deep kernel networks resides in their computational overhead. Indeed, the computational complexity of evaluating these networks scales quadratically w.r.t the cardinality of  data and this is further exaggerated in the regime of very deep networks. Existing state-of-the art solutions  mitigate this issue; for instance, authors in \cite{burges1997improving} reduce the number of kernel evaluations by approximating a heavy kernel based radial-basis function with a reduced set of kernels, and authors in \cite{mairal2014convolutional}  train  convolutional networks that best capture a particular class of invariant kernels.  Other generic solutions reduces the number of units and connection weights (and hence speedup deep-nets) using pruning and weight sharing~\cite{han2015deep,han2015deep,yu2017accelerating,han2015learning}, singular value decomposition \cite{denton2014exploiting}, regularization and sparsity~\cite{wen2016learning,changpinyo2017power,shi2017speeding,wu2017compact} as well as hardware design~\cite{han2016eie}. Our proposed solution, in this paper, is conceptually different from all the aforementioned techniques: on the one hand, our efficient kernel network design is not restricted to a specific class of kernels; it is generic and can be applied to more general classes of deep kernels (see also Eq~\ref{eq0} and Section~\ref{section3}). On the other hand, in sparsity and regularization based methods, a given (targeted) cost  may not necessarily be reached  (i.e., after solving optimization) while in our method any targeted cost, fixed a priori, can be reached  (at the expense of a some loss in precision)  thanks to the coarse-to-fine design as shown subsequently. \\

In this paper, we propose a novel coarse-to-fine framework for efficient deep kernel network evaluation.  This approach is based on a cascade of  kernel networks with a gradual increase of complexity and discrimination power. Networks in the early stages of the cascade are relatively shallow and used to reject most of   the dominant patterns (with a negligible cost) while  networks in the subsequent stages  of the cascade are more accurate (but expensive) and reserve intense computation only to the rare positive patterns. This makes the cascade very suitable for classification problems -- such as  {\it change detection} in very large satellite images -- where ``target/no-target'' classes are very imbalanced.  \\

\noindent Starting from a pretrained deep kernel network (referred to as $f$-network) which is also highly accurate and expensive, we build its surrogates (referred to as $g$-networks) with a reduced complexity using a variant of the cross-entropy criterion. The latter  minimizes the differences between the outputs of  classifiers trained on top  of the $f$ and $g$-networks. Note that the complexity of the $g$-networks (measured by the number of units and depth) is fixed a priori depending on the expected amount of computation that makes the overall evaluation cost of the cascade cheap (see also Sections~\ref{section3}, \ref{section4}). As the $g$-networks are naturally rank-deficient (i.e., their error rates are intrinsically higher than the $f$-networks), the $g$-networks are designed in order to satisfy the {\it conservation hypothesis}: the latter states that all the positive responses  of classifiers built on top of the $f$-network should be preserved by  classifiers built on top of the $g$-networks. We  implement this hypothesis  using a particular weighting scheme (of the cross-entropy criterion) that favors very small false negative rates to the detriment of an increase of false alarms. In spite of this increase of false alarms, most of these alarms (dominant patterns) are rejected in the early stages of the cascade and only a small fraction requires further processing using more expensive and accurate networks in the subsequent stages.  Note that our coarse-to-fine processing belongs to the ``$\epsilon$-lossy'' approaches that have been successfully applied  to popular problems such as face detection using hierarchies of classifiers \cite{fleuret2001coarse,amit1999computational,amit2004coarse,Li_2015_CVPR}. To the best of our knowledge, none of these solutions tackled the issue of speeding-up deep kernel networks and most of the existing solutions were dedicated to support vector machines ~\cite{kienzle2005face,sahbi2006hierarchy,romdhani2001computationally} and boosting~\cite{viola2004robust,li2004floatboost}. \\

The remainder of this paper is organized as follows; section \ref{section2} provides details about deep kernel networks while section \ref{section3} introduces the main contribution; a coarse-to-fine approach that reduces the computational complexity of these networks. Section \ref{section4} shows the efficiency and the effectiveness of our method on the challenging problem of change detection in large and high  resolution satellite imagery. Finally, section \ref{section5} concludes  the paper while providing possible extensions for a future work.  

\begin{figure}[hpb]
\begin{center} 
\includegraphics[width=9.3cm]{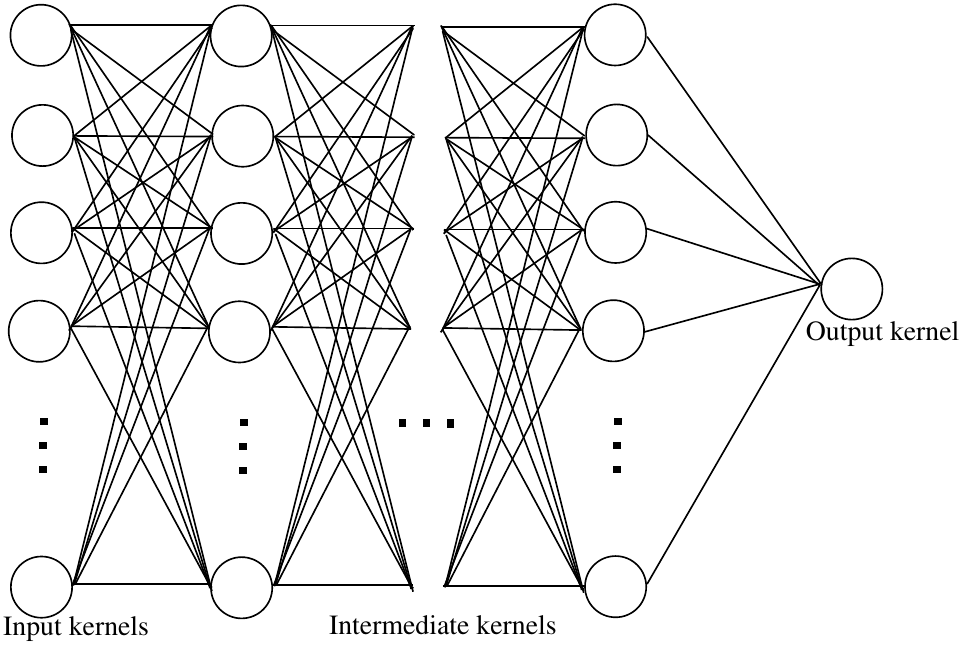}
\end{center}
\caption{This figure shows an example of a deep kernel network.}\label{fig0}
\end{figure}

\section{Deep Kernel Networks}\label{section2}

Consider a collection of \textit{$\ell$} labeled training samples ${\cal L}=\{(\mathbf{x}_i, {\bf y}_i)\}_{i=1}^\ell$, with $\mathbf{x}_i \in \mathbb{R}^d$  being a feature vector (for instance the VGG-net descriptor \cite{simonyan2014very}) and ${\bf y}_i$ its class label in $\{-1,+1\}$ and another collection of \textit{$u$} test samples ${\cal U}=\{\mathbf{x}_i\}_{i=\ell+1}^{\ell+u}$. Our goal is to jointly learn a deep multiple kernel and classifier $f$ from the labeled samples; here $f$ is an {SVM-based} classifier that predicts the class ${\bf y}_i \leftarrow \textrm{sign}[f({\bf x}_i)]$ of a given sample ${\bf x}_i \in {\cal U}$. Usual algorithms, such as shallow multiple kernel learning (MKL)~\cite{bach2004multiple}, jointly learn kernels and {SVM-based} classifiers by maximizing a margin using an EM-like optimization; as shown through this work, we consider instead a deep version of MKL which is highly efficient and effective.

\subsection{Deep Multiple Kernels} 
A kernel (denoted $\kappa$) is a symmetric function that provides a similarity
between any two given samples~\cite{cristianini2000introduction}. When positive semi-definite, $\kappa$ can be written as an inner product in a high (possibly infinite) dimensional space, via a mapping function (denoted $\phi$). Among the existing kernels, polynomial and  radial basis functions are the most studied~\cite{cristianini2000introduction}. In this work, we aim to learn an implicit mapping function that recursively characterizes a nonlinear and deep combination of multiple existing kernels.

Fig. \ref{fig0} shows our deep kernel network with $L$ layers. For each layer $l$ and its associated unit $p$, a kernel domain $\big\{ {\kappa}_{p}^{(l)}(\cdot,\cdot) \big\}$ is recursively defined as
\begin{equation}\label{eq0}
{\kappa}_{p}^{(l)}(\cdot,\cdot) = h \big(\sum_q {\bf w}_{q,p}^{(l-1)} \
\kappa_q^{(l-1)}(\cdot,\cdot) \big),
\end{equation}
\noindent where $h$ is a nonlinear activation function\footnote{In all this work, we use the Rectified Linear Unit (ReLU); the latter is defined as $h(\x) =\max(0,\x)$.}. In the above equation, $q \in \{1,\dots,n_{l-1}\}$, $n_{l-1}$ is the number of units in layer $(l-1)$ and $\{\mathbf{w}^{(l-1)}_{q,p}\}_q$ are the (learned) weights associated to kernel ${\kappa}_{p}^{(l)}$. In particular, $\{{\kappa}_{p}^{(1)}\}_p$ are the input kernels including Gaussian, etc. When $L=2$, the architecture is shallow, and it is equivalent to the nonlinear version of MKL (see for instance Zhuang et al.~\cite{zhuang2011two}). For larger values of $L$, the network becomes deep. 

\noindent For any given pair of samples, a vector containing the values of different standard (elementary) kernels on this pair is evaluated and considered as an input to our deep network. These elementary kernel values are then forwarded to the subsequent intermediate layer resulting into $n_2$ multiple kernels through the nonlinear combination of the previous layer, etc. The final kernel is a highly nonlinear combination of elementary kernels. \\

 Note that with this setting, deep kernel network evaluation is inductive, and the computation feasible on any new pairs of samples. Note also that the deep kernel network in essence is a multi-layer perceptron (MLP), with nonlinear activation functions. The difference is that the last layer is not designed for classification, rather than to deliver a similarity value. However, we can use the classical backpropagation algorithm specific for MLP to optimize the weights in the deep kernel network. Let $J$ denotes an objective function associated to our classification problem. More details, about choice of $J$, are discussed in Section~\ref{section3}. We assume that  the computation of gradients of the objective function $J$ w.r.t the output kernel $\kappa^{(L)}_1$ (i.e.~$\frac{\partial J}{\partial \kappa^{(L)}_1(.,.)}$) is tractable. According to the chain rule, the corresponding gradients w.r.t coefficients $\mathbf{w}$ are computed, and then used to update these weights using gradient descent.

\section{Coarse-to-Fine Deep Kernel Networks}\label{section3}

\begin{figure}[hpb]
\begin{center} 
\includegraphics[width=12.5cm]{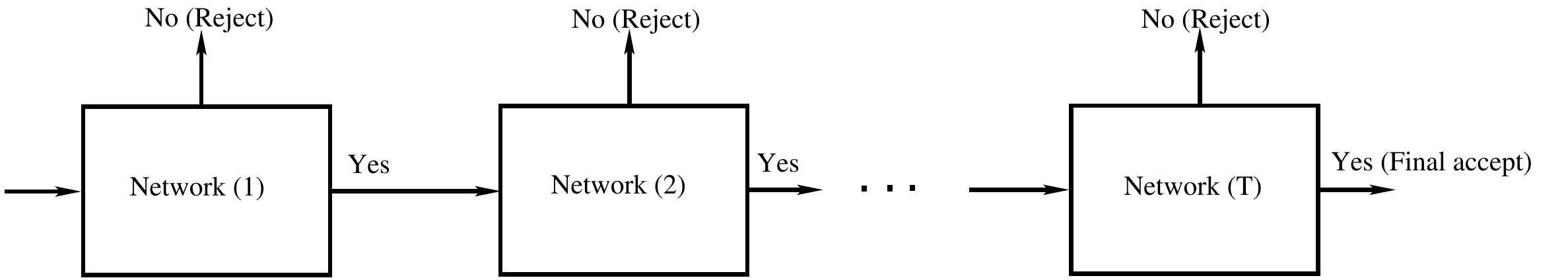}
\end{center}
\caption{This figure shows the cascade of $T$ classifiers and $g$-(kernel)-networks (as shown in experiments,  $T$ is set to 6).}\label{cascade} 
\end{figure}

Let $f$ be an SVM classifier trained on top of the deep kernel $\kappa_1^{(L)}$; in what follows, $\kappa_1^{(L)}$ is simply rewritten  as $\kappa_f$. In practice, the depth of this deep kernel network (and also the number of its units) should be sufficiently large in order to optimize the generalization performance of $f$ (see~\cite{jiu2017nonlinear}). However, deep kernel networks may affect the computational efficiency of $f$ as the evaluation cost of the underlying deep kernel $\kappa_f$ becomes extremely prohibitive; particularly on limited hardware resources. \\
\noindent Our goal is to make the evaluation cost of these kernel networks cheap by reducing their complexity while maintaining their high accuracy. As shown through this paper, this is achieved using a well optimized cascade of deep networks (and classifiers) that quickly rejects simple patterns which belong to the {\it dominant} class while reserving intense computation only to the {\it rare} targeted class (see Fig.~\ref{cascade}).

\subsection{The $f$-network vs. the $g$-networks}
In what follows, the $f$-network refers to the original deep kernel network while its $g$-variant corresponds to its simplified version. Considering a pretrained classifier $f$ (and its associated $f$-network), building a single monolithic classifier $g$ (on top of a reduced complexity $g$-network) -- which is strictly equivalent to $f$ -- is clearly out of reach; at least because the representational power\footnote{The representational power of a deep kernel network is closely related to the number of its parameters. As the number of parameters increases, the maximum number of samples that can be accurately shattered (whatever their labeling) becomes larger.} of the $f$-network is higher than the $g$-network. \\
\indent In order reduce the computational complexity of evaluating $f$, we proceed differently. We consider a {\it coarse-to-fine} cascade of classifiers $\{g_t\}_{t=1}^T$; any given $g_t$ is a simplified instance of $f$. More precisely, $\{g_t\}_t$ are associated to deep kernel networks with increasing complexities (measured by the depth and number of units). Test patterns are fed to this cascade and classified in a coarse-to-fine way; a test pattern $\x$ is declared as positive iff all the classifiers $\{g_t\}_{t=1}^T$ answer positively, while $\x$ is quickly rejected (as negative) if one of the classifiers  $\{g_t\}_t$ answers negatively. As the negative class is dominant, the overall evaluation cost of the cascade is dominated by the cost to reject the negative patterns. Hence, the high efficiency of the cascade is dependent on the ability of the classifiers (in the early stages) to reject negative patterns quickly while maintaining the positive scores of the $f$-classifier (i.e., classifier at the final stage).  The latter property is written as 
$$ \forall \x, \forall t=1,\dots,T, \ \ \  f(\x) > 0 \implies  g_t(\x)> 0$$
\noindent this property is referred to as the {\it conservation hypothesis} which means that all the $g$-classifiers should be implemented in order to preserve all the positive answers of the initial $f$-classifier. \\

We want $\{g_t\}_t$ to be both efficient and with equivalent error rates compared  to $f$. Again, as the representational power of a $g$-network (w.r.t its associated $f$-network) is limited, seeking to obtain the same false positive (FP) and false negative (FN) rates (w.r.t $f$) is clearly out of reach. Hence, we seek to make the FN rate of the $g$ and $f$ networks as close as possible (by implementing the conservation hypothesis) to the detriment of an increase of the FP rate of the $g$-network. With this setting, patterns classified as positive in the early stage of the cascade will further be processed through the subsequent stages and only those belonging (or resembling) to the targeted class will reach the final stage of the cascade; hence the overall evaluation cost of the cascade will be dominated by the cost to reject negative patterns using very cheap $g$-networks.

\subsection{Learning cheap $g$-networks} 

Given a classifier $f$, we aim to design its surrogate cascade classifiers $\{g_t\}_t$ with fixed complexities (again defined by the depth and number of units; see Section~\ref{section4}). For a fixed stage $t$ in the cascade, we rewrite its  classifier $g_t$ simply as $g$ and the output of its $g$-network as $\kappa_g$. We propose to find the parameters of the $g$-network by minimizing the following loss 
\begin{equation}
\begin{array}{lll}
\displaystyle   \min_{\bf w} & \displaystyle   \beta^- \sum_{i=1}^\ell  1_{\{f(\x_i) \leq  0\}}  \ \frac{1}{1+\exp(-\gamma g(\x_i))} \\
                    &  \ \ \ \ \ \ \ \ \ \ \ \ \ \  +  \  \displaystyle \beta^+ \sum_{i=1}^\ell 1_{\{f(\x_i) > 0\}} \ \frac{\exp(-\gamma g(\x_i))}{1+\exp(-\gamma g(\x_i))}
\end{array}\label{eq1}
\end{equation}

\noindent here $\bf w$ are the weights of the $g$-network (as defined in Eq.~\ref{eq0}), $\small f(\x) =  \sum_i \alpha_i^f \y_i \kappa_f (\x,\x_i) + b_f$ is assumed pretrained and $g(\x) =  \sum_i \alpha^g_i f(\x_i) \kappa_g(\x,\x_i) + b_g$ with $(\{\alpha^g_i\}_i,b_g)$ trained as shown in Section~\ref{section00}. In the above objective function, $\beta^+, \beta^- \geq 0$ and $\gamma \geq 0$ is the steepness of the logistic function ($1/(1+\exp(-\gamma g(\x) )$) set in practice to very large values, so this function acts as  a smooth differentiable variant of the 0-1 loss. When $\beta^+ \gg \beta^-$, this re-balances the FP and FN rates in a way that favors very small FN of $g$ w.r.t $f$ (i.e., it makes it possible to implement the conservation hypothesis) to the detriment of an increase of the FP rate.\\
\noindent Note that this criterion is a variant of the cross-entropy loss; it seeks to reduce the number of contradictory outputs from the classifiers $f$ and $g$. Note also that this criterion does not require labeled data provided that the classifier $f$ (and its $f$-network) are already pretrained\footnote{Training the $f$-classifier and its $f$-network requires minor updates of the objective function~\ref{eq1} (and also \ref{equa:hingeloss}); only the indicator function terms are set according to the actual labels of data in $\{\x\}$ (i.e., $1_{\{f(\x)> 0\}}$ is replaced by  $1_{\{\y=+1\}}$ and  $1_{\{f(\x)\leq 0\}}$ by  $1_{\{\y=-1\}}$).}; as shown in Section~\ref{section00}, the outputs of the classifier $f$ are used as reference labels, so this optimization framework could benefit from very large unlabeled sets in order make the estimate of the $g$-classifiers (and their $g$-networks) more accurate.

\subsection{Optimization}\label{section00}

The goal is also to learn the SVM parameters $(\{\alpha_i^g\}_i,b_g)$ on top of the current estimate of $\kappa_g$. For that purpose, we use the forward information $\kappa_g(\cdot,\cdot)$ from the learned $g$-network in order to build a binary classifier $g$ by minimizing a global hinge loss and a regularization term
\begin{equation}
\min_{g} \ \ C \ \sum_{i=1}^\ell \max \big(0, 1 - f(\mathbf{x}_i) g(\mathbf{x}_i)\big) + \frac{1}{2} \big\|g \big\|_{\mathcal{H}}^2,
	\label{equa:hingeloss}
\end{equation}
\noindent here  $C \geq 0$ controls the tradeoff between regularization and empirical error. According to the representer theorem \cite{cristianini2000introduction}, the dual form of Eq.~\ref{equa:hingeloss} can be rewritten as
{
\begin{align}
\max_{\alpha^g} & \sum_{i=1}^\ell \alpha_i^g 
-\frac{1}{2}  \sum_{i,j=1}^\ell \alpha_i^g \alpha_j^g f(\x_i)  f(\x_j) \   
\kappa_g({\bf x}_i,{\bf x}_j) 
\nonumber     \\
 &\textrm{s.t.} \quad 0 \leq \alpha_i^g \leq C, \  \sum_{i=1}^\ell \alpha_i^g f(\x_i) = 0,
\label{equa:maxdualproblem} 
\end{align}
}

\noindent following the KKT conditions~\cite{burges1998tutorial}, $b_g$ is a shift that guarantees the equilibrium constraint (in Eq.~\ref{equa:maxdualproblem}). \\

The objective functions~\ref{eq1} and \ref{equa:maxdualproblem} are optimized w.r.t two parameters: respectively weights $\mathbf{w}$ of the $g$-network and $(\alpha^g,b_g)$ of classifier $g$. Alternating optimization strategy is adopted, i.e., we fix $\mathbf{w}$ to optimize $(\alpha^g,b_g)$, and then vice-versa. At each iteration, when $\mathbf{w}$ is fixed, $\kappa_g(.,.)$ is also fixed, and $(\alpha^g,b_g)$ are optimized using an SVM solver (LIBSVM in practice~\cite{chang2011libsvm}). When $(\alpha^g,b_g)$ are fixed, the gradient of Eq.~\ref{eq1} w.r.t the output $k_g$ (of the $g$-network) is evaluated, and a round of backpropagation is achieved and $\mathbf{w}$ is accordingly updated using gradient descent. The iterative procedure (shown in algorithm~\ref{algo:supervised}) continues until convergence or when a maximum number of iterations is reached.

\begin{algorithm}[hbpt]
\caption{Deep Kernel Network Learning}
\label{algo:supervised}
\BlankLine
\KwIn{Initial $\mathbf{w}^{(l)}(l=1, \ldots, L-1)$}
\KwOut{Optimal $\mathbf{w}^{(l)}(l=1, \ldots, L-1)$, $\alpha^g$, $b_g$}
\Repeat{Convergence}
{
	Fix $\mathbf{w}$, compute the output kernel $\kappa_g({\bf x}_i,{\bf
x}_j), \forall i, j \in {1, \ldots \ell}$\;
    $\alpha^g, \beta_g$ are learned by the LIBSVM solver\;
	Fix $\alpha^g, b_g$, compute the gradient of \ref{eq1} w.r.t
$\{\kappa_g(\mathbf{x}_i, \mathbf{x}_j)\}_{ij}$\;
	Update $\mathbf{w}$ (and hence $\kappa_g$) using backpropagation and gradient descent;
}
\end{algorithm}
\section{Experiments}\label{section4}

\noindent {\bf Datasets and Task:} we evaluate the performance of our proposed method on the  challenging task of satellite image change detection. The goal is  to find instances of relevant changes into a given scene acquired at instance $t_1$ with respect to the same scene taken at instant $t_0 < t_1$; these acquisitions (at instants $t_0$,  $t_1$) are referred to as {\it reference} and {\it test} images respectively. This task is known to be very challenging due to the difficulty to characterize relevant changes (appearance or disappearance of objects\footnote{This can be any object so there is no supervision or a priori knowledge about what object may appear or disappear into a given scene.}) from  irrelevant ones (such as the presence of cars, clouds, etc.), and it is also  very time demanding as the amount of data to process on large geographic areas is extremely large. Indeed, with the spread of remote sensors and unmanned aerial vehicles (UAV), and in the particular important scenario of damage assessment after natural hazards (such as tornadoes, earth quakes, etc.), it is crucial to achieve automatic change detection very  promptly in order to organize and prioritize rescue operations; that's why one should use very accurate learning and classification algorithms (such as deep networks)  while being able to process large amount of data efficiently.  \\

\begin{figure*}[hpbt]
\begin{center} 
\includegraphics[width=5.6cm]{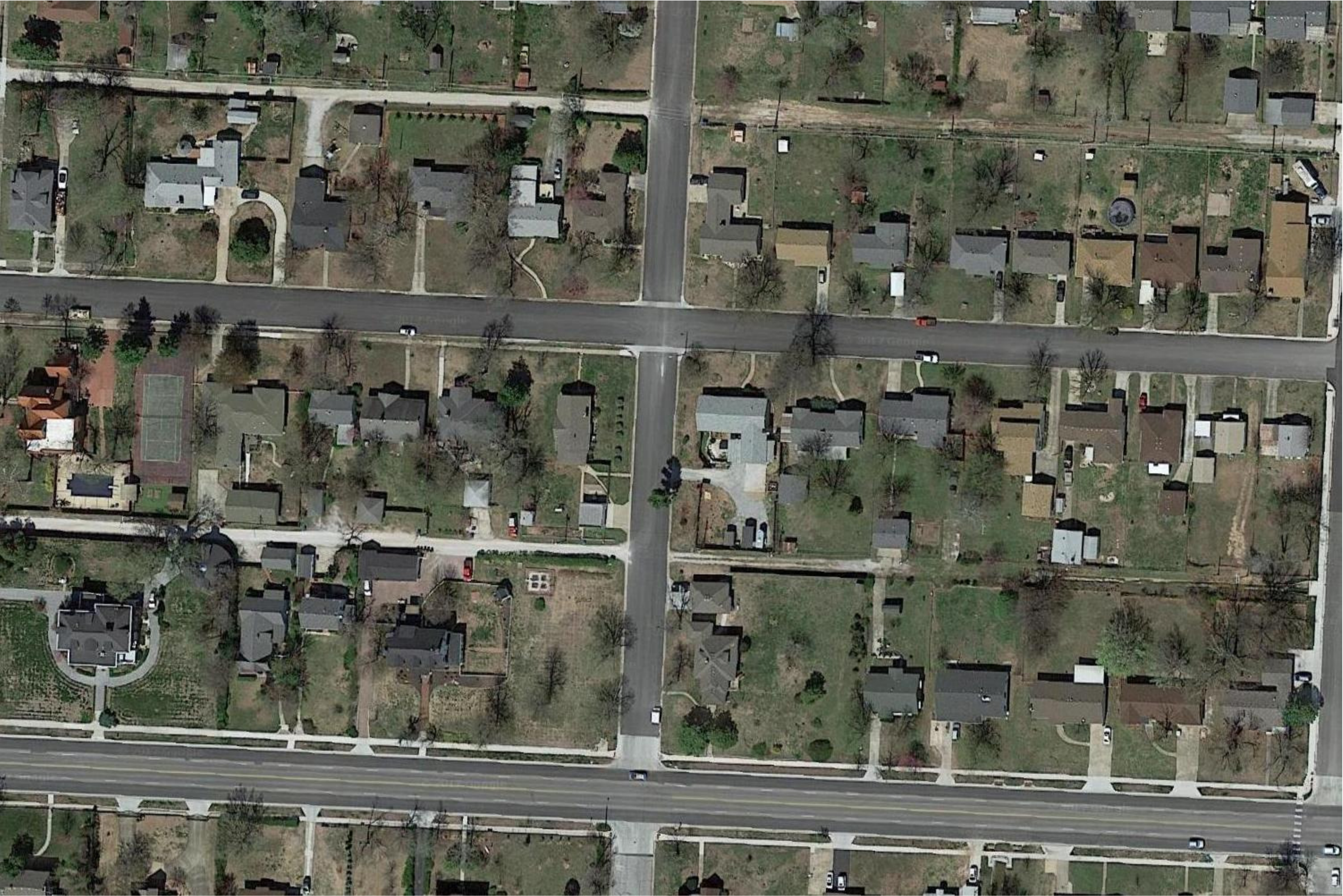}\hspace{0.2cm}\includegraphics[width=5.6cm]{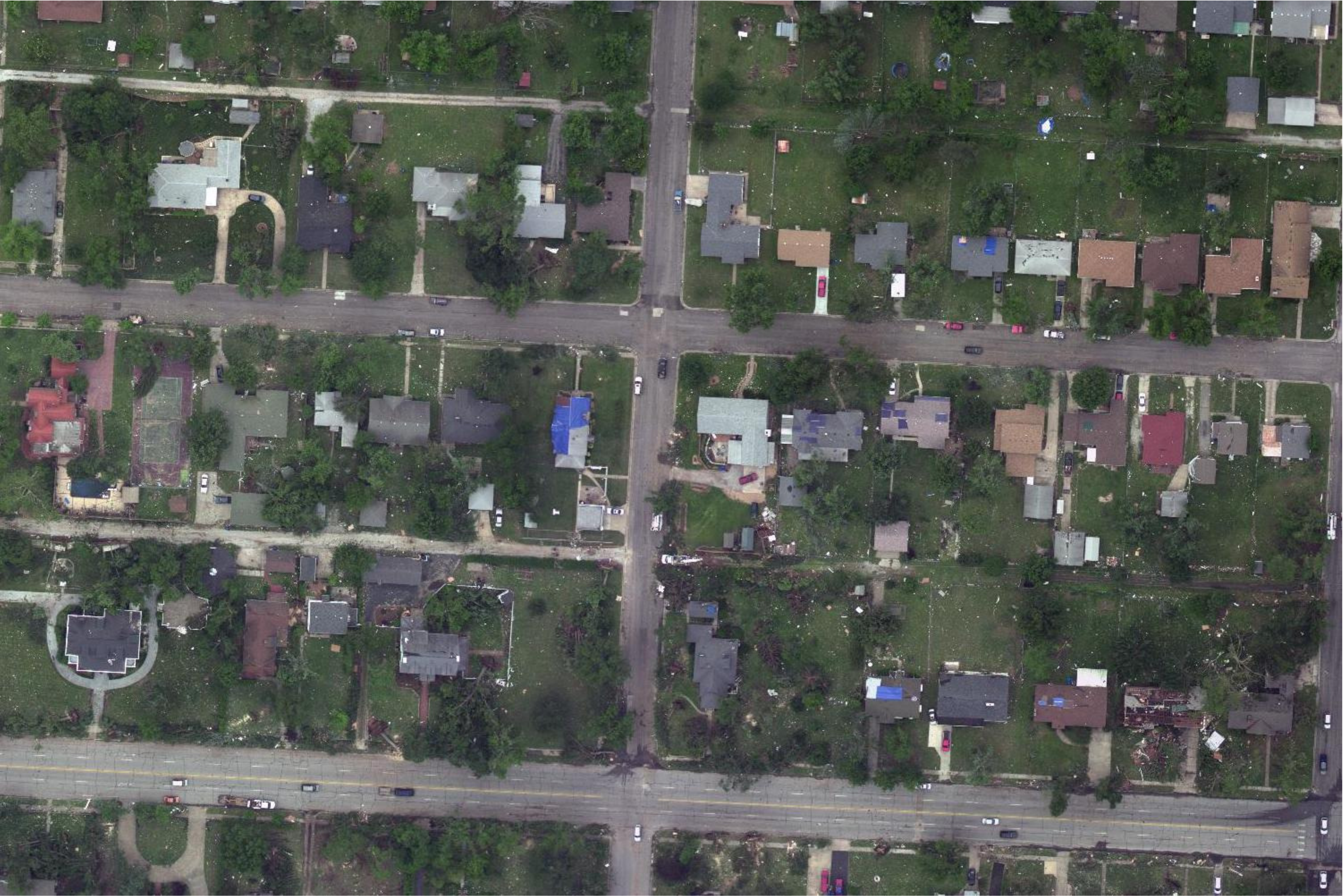}\hspace{0.2cm}\includegraphics[width=5.6cm]{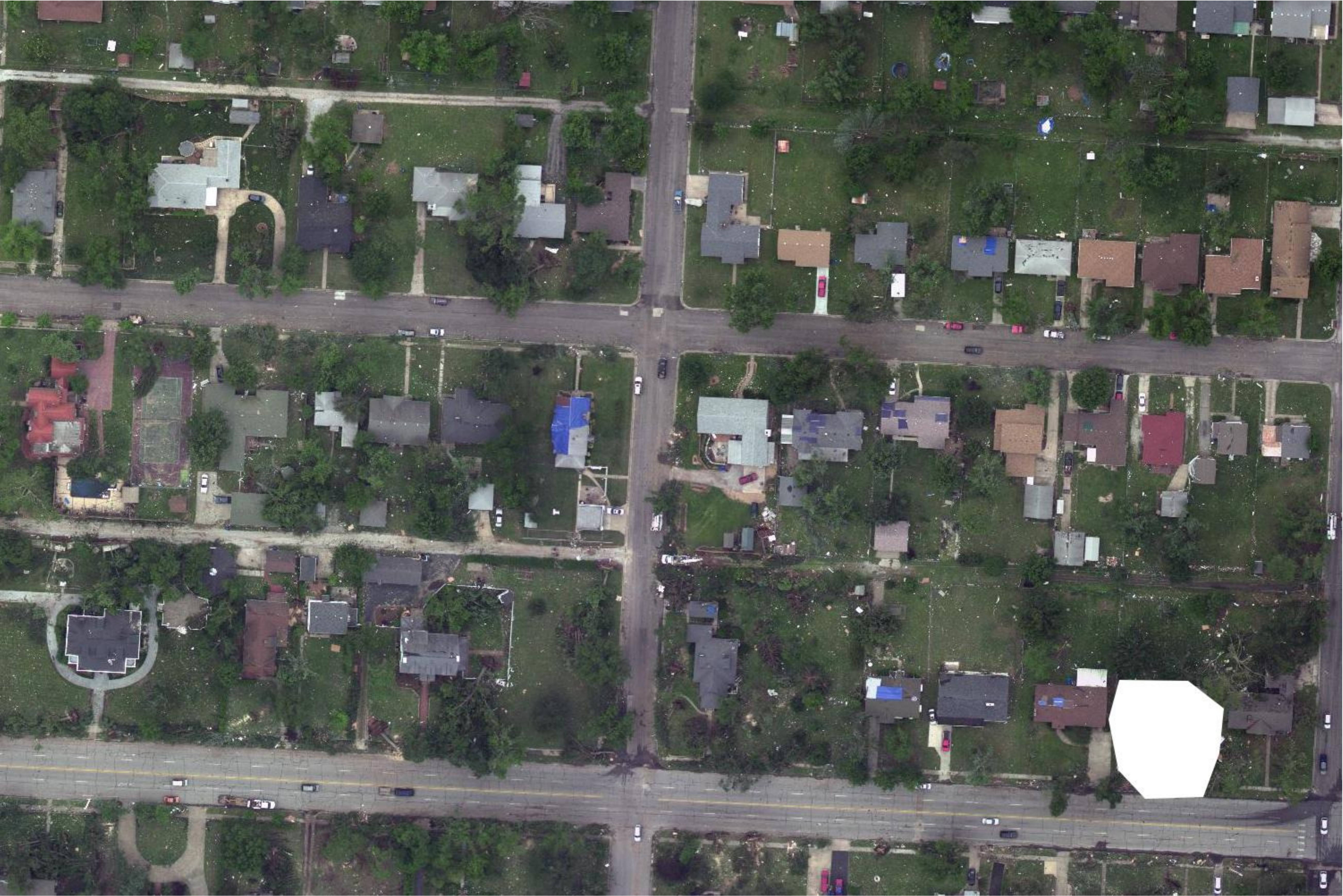}
\end{center}
\caption{This figure shows an example of a district hit by a tornado: (left) reference image, (middle) test image and (right) a mask of (a hand-labeled) change detection ground-truth shown in white (severe house damage).}\label{fig2}
\end{figure*}

\indent Considering this scenario, we use a database ${\cal L}\cup {\cal U}$ of $680928$ non overlapping patch pairs (of $30\times30$ pixels in RGB) taken from six registered (reference and test) GeoEye-1 satellite images (of $9850\times10400$ pixels each).  
These images cover a very large area -- of about $20 \times 20 \ {\bf km}^2$  -- around Joplin (Missouri; see an example of a district from this area in Fig.~\ref{fig2}) and show many changes after tornadoes that happen in may 2011 (building destruction, etc.)  and no-changes (including irrelevant ones such as car appearance/disappearance, etc.).
Each patch pair (in reference and test images) is encoded  with $4096$ coefficients corresponding to the difference between the outputs of the $4096$-dimensional-layer (of the pretrained VGG-net~\cite{simonyan2014very}) on the reference and test patches. A given patch pair, denoted ${\bf x}$ (with ${\bf x} \in   {\cal U}$), is declared as a ``change'' or ``no-change'' depending on the scores of the trained SVM classifiers.  \\

\noindent {\bf Evaluation measures:} in order to evaluate the performances  of our change detection classifiers, we use the following evaluation measures  
\begin{itemize} 
\item False alarm  (FA) and detection rate (DR): the former is the fraction of ground truth "no-changes" which are declared as positive while the latter is the fraction of ground truth "changes" which are correctly classified as positive.  Smaller  FA and higher  DR imply better performances.
\item  The equal error rate (EER) : this is defined as the average  between FA and (1-DR). EER is the balanced generalization error that equally weights errors in ``change'' and ``no-change'' classes. Smaller EER implies better performance. 
\item The relative (rFA) and the conservation rate (cons): these two measures are similar to FA and DR respectively with the only difference being the ground truth which is taken from the $f$-classifier instead of the original ground truth. 
\end{itemize}
 \noindent all these measures are evaluated on the unlabeled data  in ${\cal U}$. \\
 
\noindent {\bf Pretraining the $f$-network:}  this kernel network is  fully connected and has 8 layers with 128 units per layer  excepting  the output layer which has a single unit;   these layers  consist of convolutional units followed by rectified linear units (ReLU).  The 128 input kernels correspond to the values of the Gaussian similarity function\footnote{with a scale factor set to the average distance between the VGG descriptors of data and their neighbors.} evaluated on 128   disjoint chunks taken from  the $4096$ VGG dimensions;  hence,  each chunk   has $4096/128=32$ dimensions. \\

\begin{figure*}[hpbt]
\begin{center} 
\includegraphics[width=5.8cm]{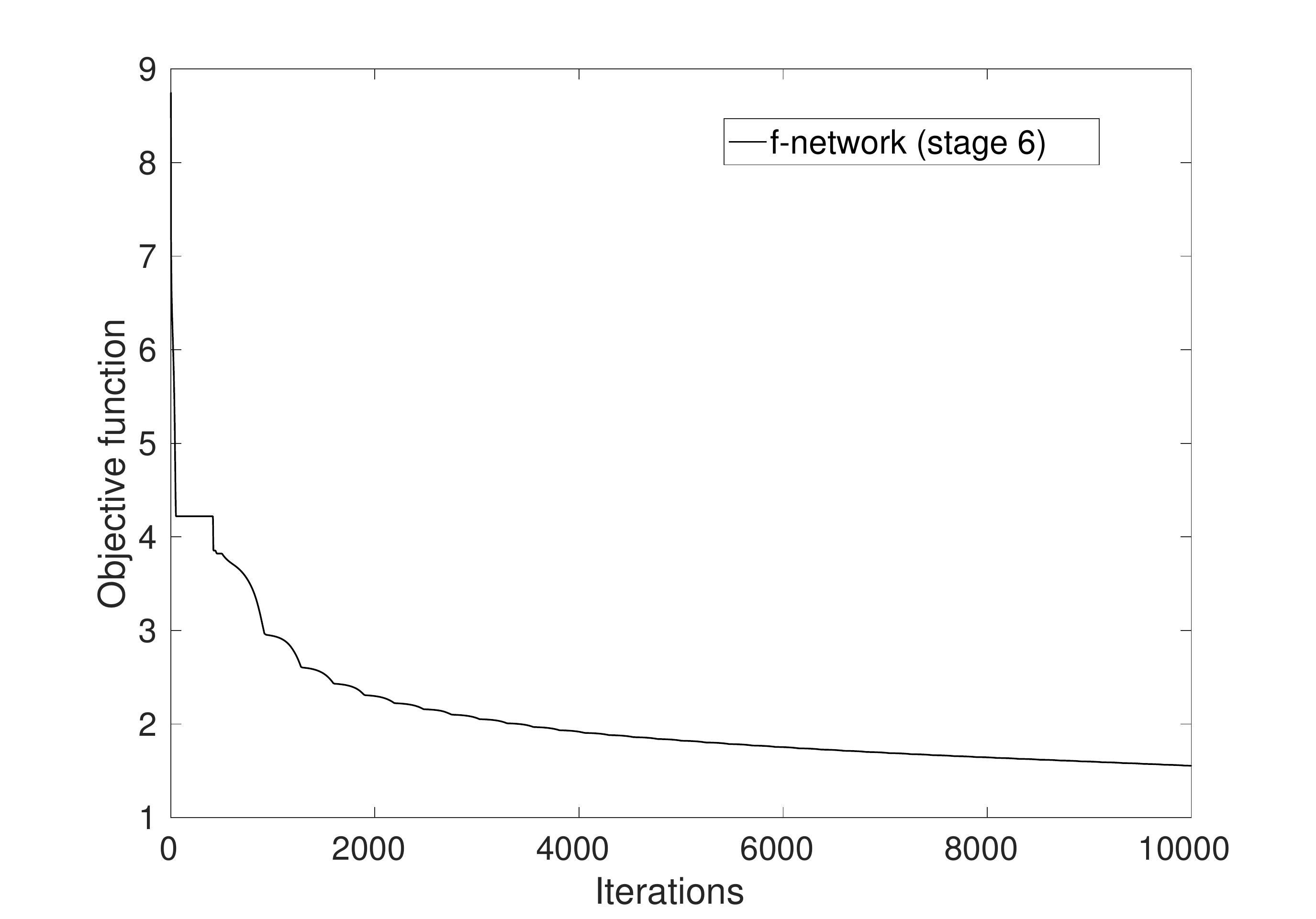}\includegraphics[width=5.8cm]{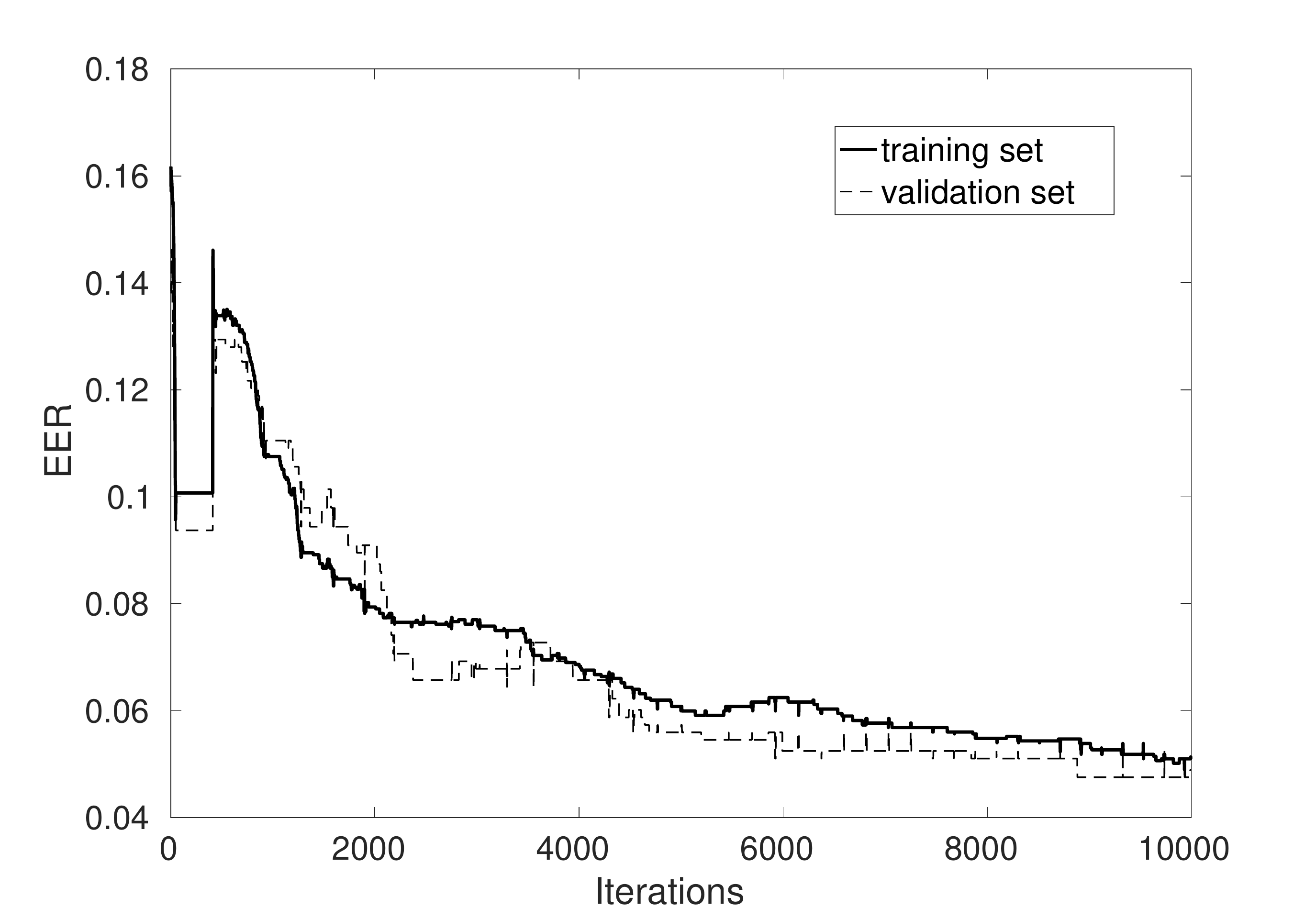}\includegraphics[width=5.8cm]{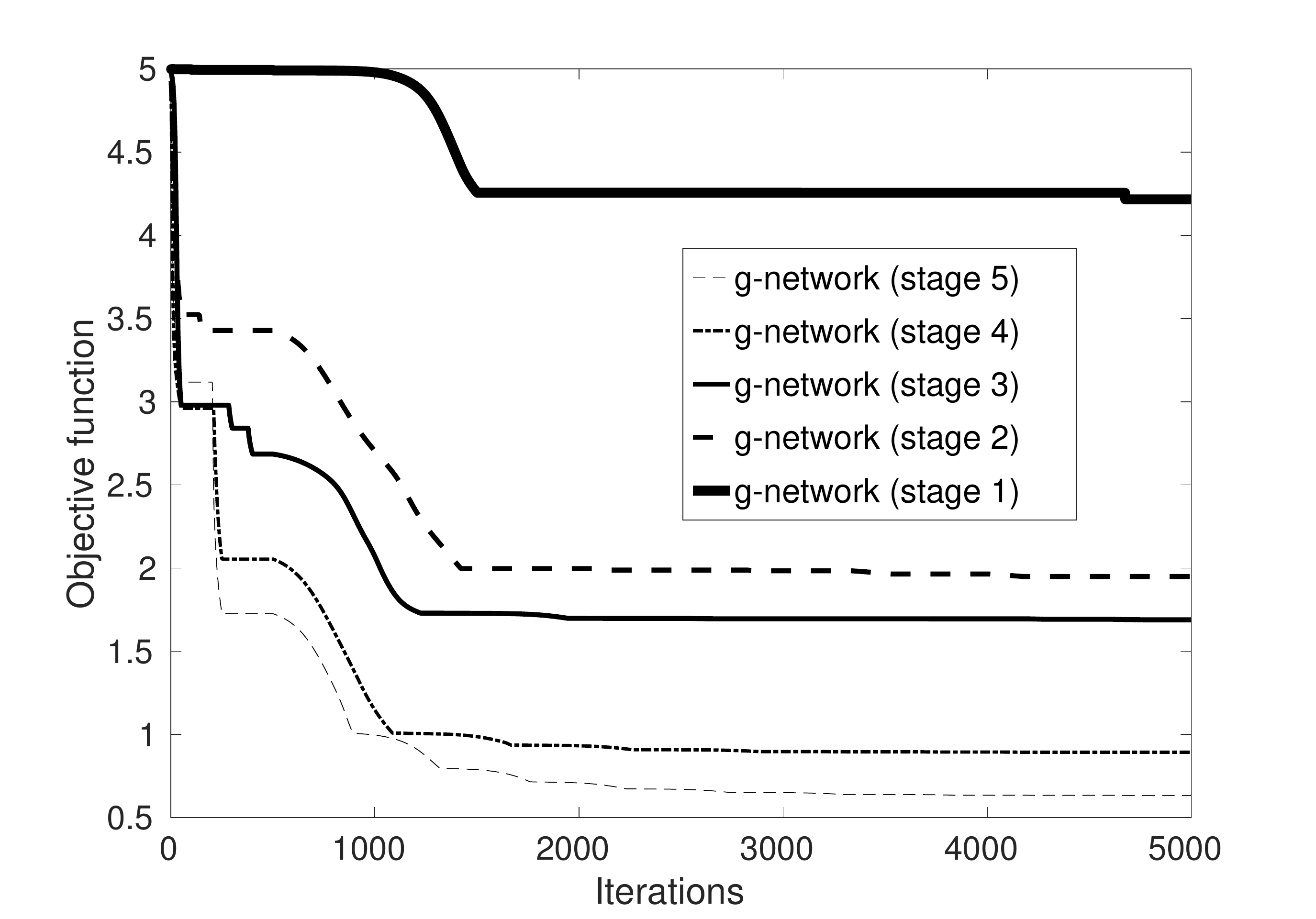}
\end{center}
\caption{This figure shows (left)  the evolution of the objective function~\ref{eq1} used to train the $f$-network w.r.t different iterations of optimization, (middle) the EER of the underlying $f$-classifier on training and validation sets, (right) the evolution of the objective functions~\ref{eq1} (used to train the $g$-networks) w.r.t different iterations of optimization.}\label{fig1}
\end{figure*}

 In order to train the parameters of this $f$-network we consider a random subset $\cal L$ including $3000$ patch pairs from the original set of $680928$ patch pairs; $2/3$ of $\cal L$ are split into 10 mini-batches for training (i.e. to minimize the objective function~\ref{eq1})  while the remaining $1/3$ is used as a validation set.  The test error is reported on all the remaining $(680928-3000)$ patch pairs. The weights of the $f$-network, set initially as flat, are updated using stochastic gradient descent (SGD) and back-propagation as shown in algorithm (1); the latter is run iteratively a max number of epochs (set in practice to $10000$)  in order to obtain convergence (see Fig.~\ref{fig1}, left) and this is observed in less than one hour on a standard PC with a 3Ghz  CPU. When training the $f$-network, $\beta^+$,  $\beta^-$  are set proportional to $\frac{1}{|\{ \y_i =+1\}_i|}$, $\frac{1}{|\{ \y_i =-1\}_i|}$ respectively and the step-size of SGD (denoted as $\nu$) is set iteratively inversely proportional to  the speed of change of the objective function~\ref{eq1}; when this speed increases (resp. decreases), $\nu$  decreases as  $\nu \leftarrow \nu \times 0.99$ (resp.   increases as $\nu \leftarrow  \nu / 0.99$). Table~\ref{tab1} and Fig.~(\ref{fig1}, left/middle) show the evolution of the objective function as well as the equal error rates of the $f$-network w.r.t different iterations of the optimization process shown in algorithm (1). \\

\begin{table}
\begin{center}
 \resizebox{0.7\columnwidth}{!}{
\begin{tabular}{c||c|c}
 & \% EER (Training) & \% EER (Validation)  \\ 
  \hline
  \hline
Weights (initialization) &  16.18  &  13.85 \\ 
Weights (at convergence)  & 05.14  &  04.90  
\end{tabular}}
\vspace{0.1cm}
\caption{This table shows the EER of the $f$-networks at initialization and at the end of the iterative process.}\label{tab1}
\end{center}
\end{table} 

\noindent {\bf Training the cascade of the $g$-networks:}  in order to build the $g$-networks of our  cascade, we consider the following   architectures (complexities)  
\begin{itemize} 
\item  {\bf Stage 1:}   this kernel network is  fully connected and has 4 layers with 2 units per layer;  these layers  consist of convolutional units followed by rectified linear units (ReLU) excepting  the output layer which has a single convolutional unit followed by a ReLU.  
\item  {\bf Stage 2:}   this kernel is similar to the previous one except that the number of units  per layer is 8.   
\item  {\bf Stage 3:} the only difference w.r.t stage 2 is the number of layers which is now set to 6.  
\item  {\bf Stage 4:} the only difference w.r.t stage 3 is the number of units per layer which is now set to 32.  
\item  {\bf Stage 5:}  this kernel is similar to the previous one except that the number of units per layer is 64.  
\item  {\bf Stage 6:}  this network is exactly the pretrained $f$-network.\\  
\end{itemize} 

Similarly to the $f$-network, we use the same splits of data (into training, validation and test sets, etc.) in order to learn the parameters of the $g$-networks.  These parameters, set initially as flat, are again updated using SGD and back propagation as shown in algorithm (1); in these experiments, the  max number of epochs is now set to  $5000$  as the number of parameters in the $g$-networks is smaller compared to the $f$-network. With this setting,  convergence is observed  in less than 30mins using the same hardware configuration. In all these experiments, the step-size of SGD is set as described earlier while $\beta^+$,  $\beta^-$  are now set proportional to   $\frac{0.99}{|\{ f(\x_i)> 0\}_i|}$, $\frac{0.01}{|\{ f(\x_i)\leq  0\}_i|}$  in order to implement the conservation hypothesis. \\

\begin{figure*}[hpbt]
\begin{center} 
  \includegraphics[width=5.6cm]{figures/figure8}\hspace{0.2cm}\includegraphics[width=5.6cm]{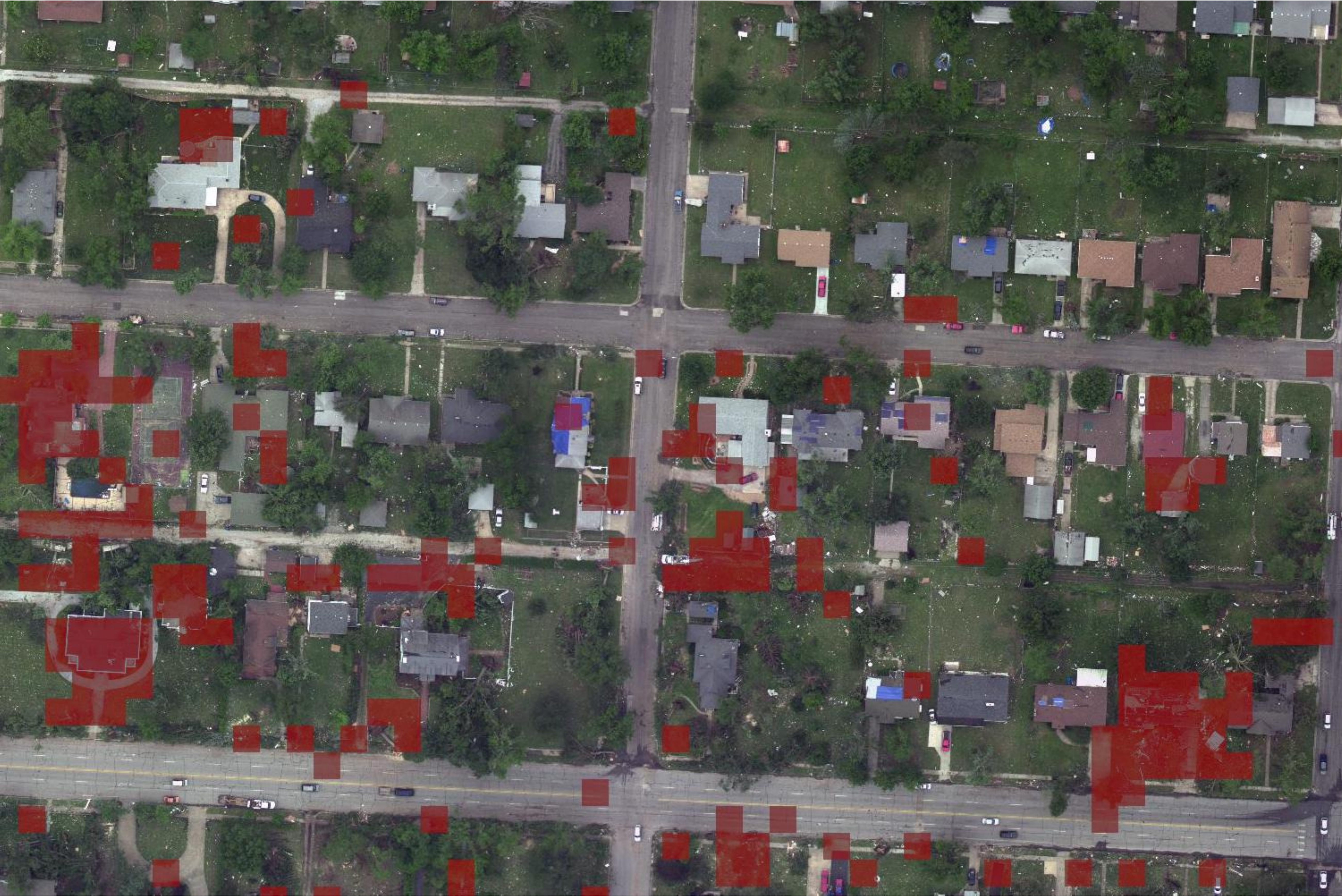}\hspace{0.2cm}\includegraphics[width=5.6cm]{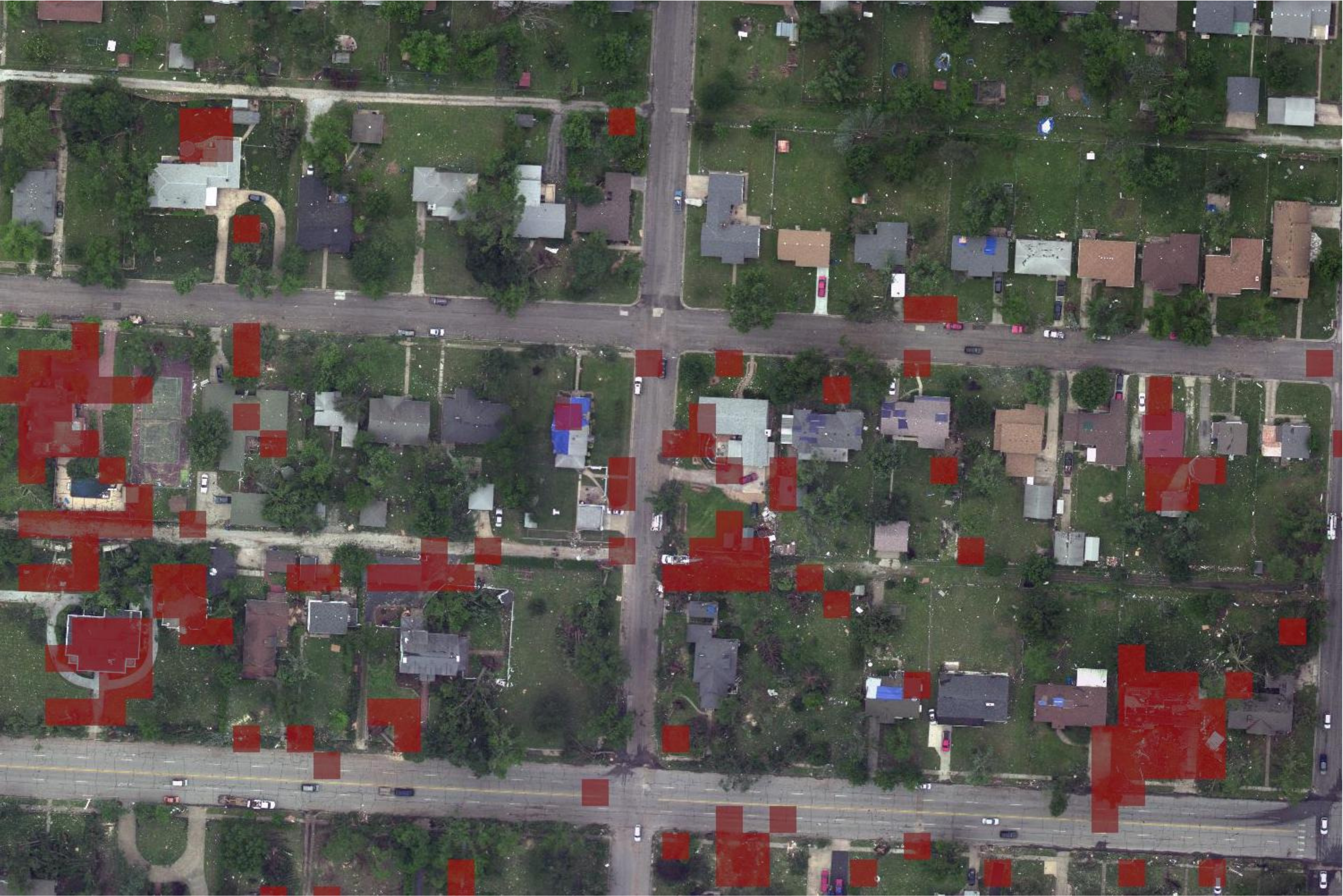}\\
(stage 1)  \hspace{5cm}  (stage 2)  \hspace{5cm}  (stage 3)  
\vspace{0.2cm}
\includegraphics[width=5.6cm]{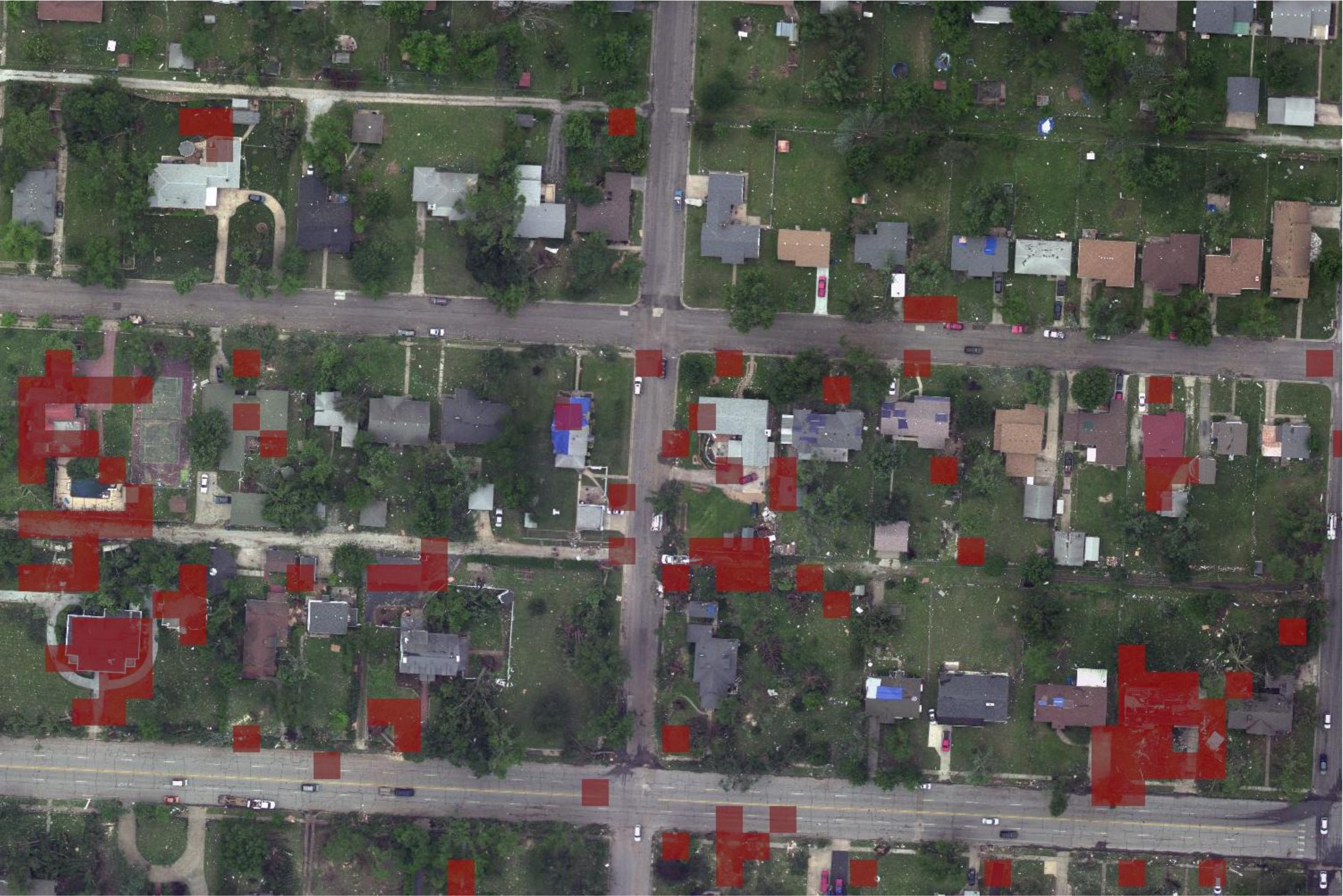}\hspace{0.2cm}\includegraphics[width=5.6cm]{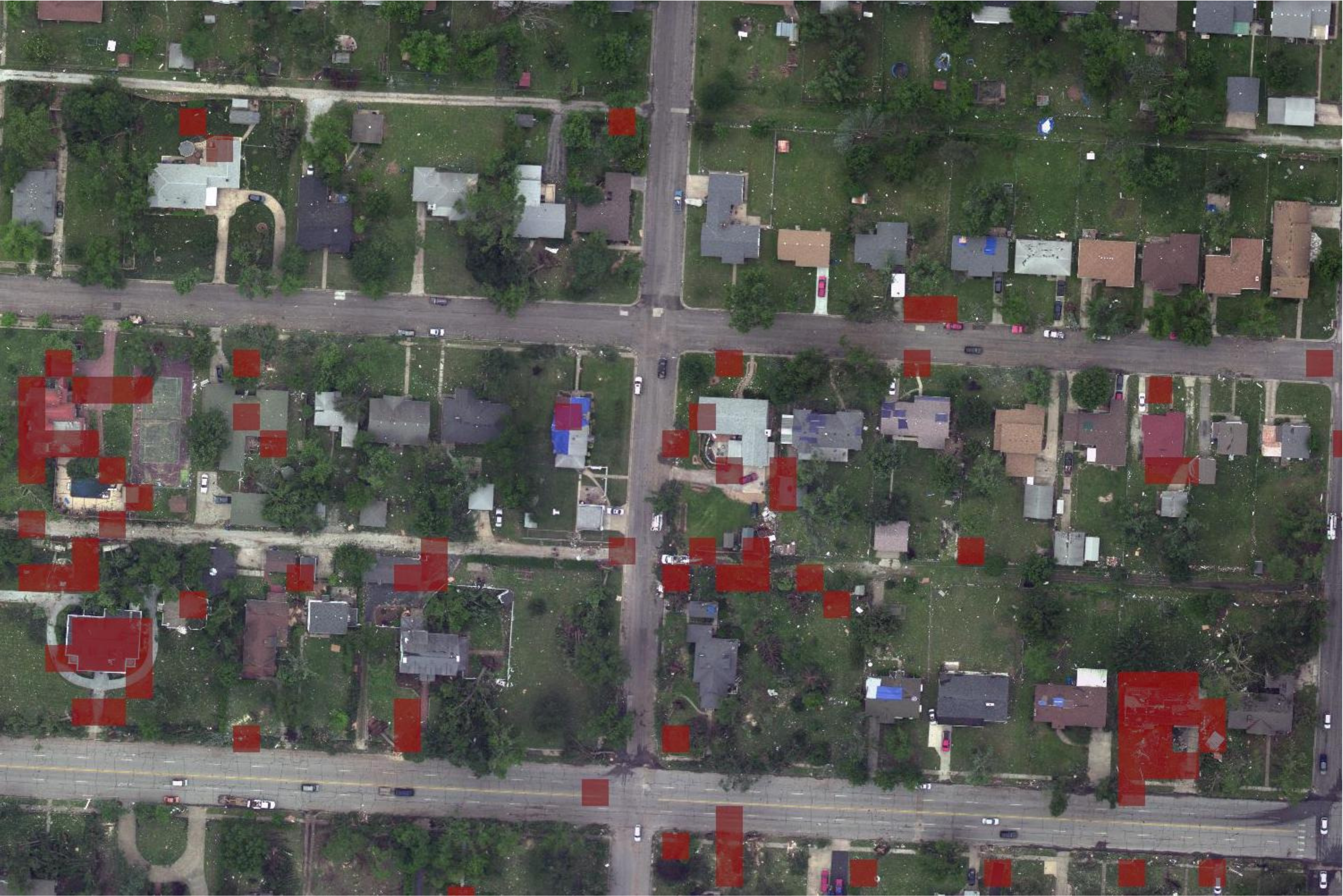}\hspace{0.2cm}\includegraphics[width=5.6cm]{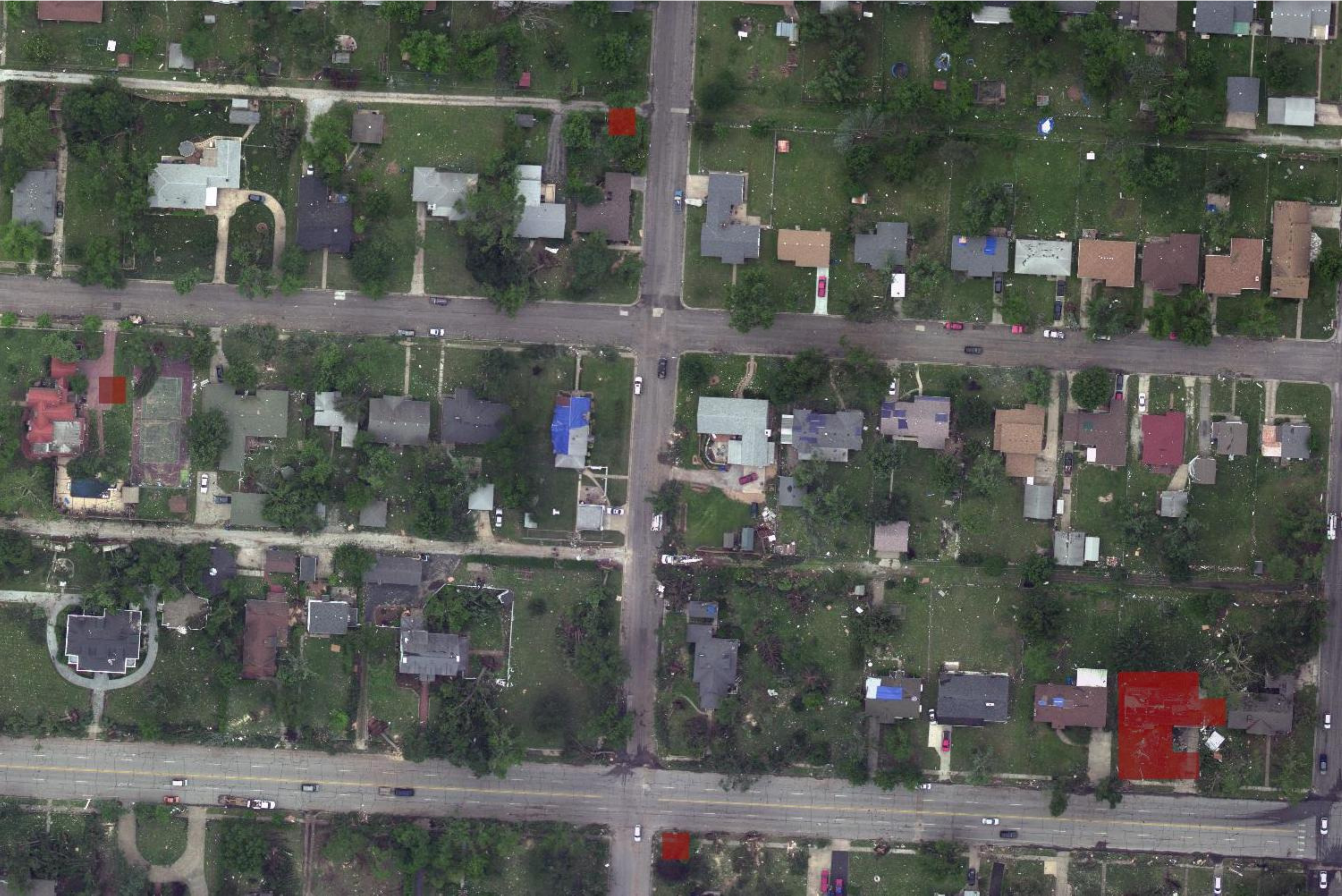} 
(stage 4)  \hspace{5cm}  (stage 5)  \hspace{5cm}  (stage 6)  
\end{center}
\caption{This figure shows the evolution of detections through the different stages of the cascade; as we go through different stages of the cascade the {\it global} number of false alarms decreases (in contrast to the setting of table 2, a classifier, at a given stage, is applied only to the patterns declared as positive by the preceding stages).} \label{fig3}
\end{figure*}

\begin{figure*}[hpbt]
\begin{center} 
\includegraphics[width=8cm]{figures/figure4}\hspace{0.2cm}\includegraphics[width=8cm]{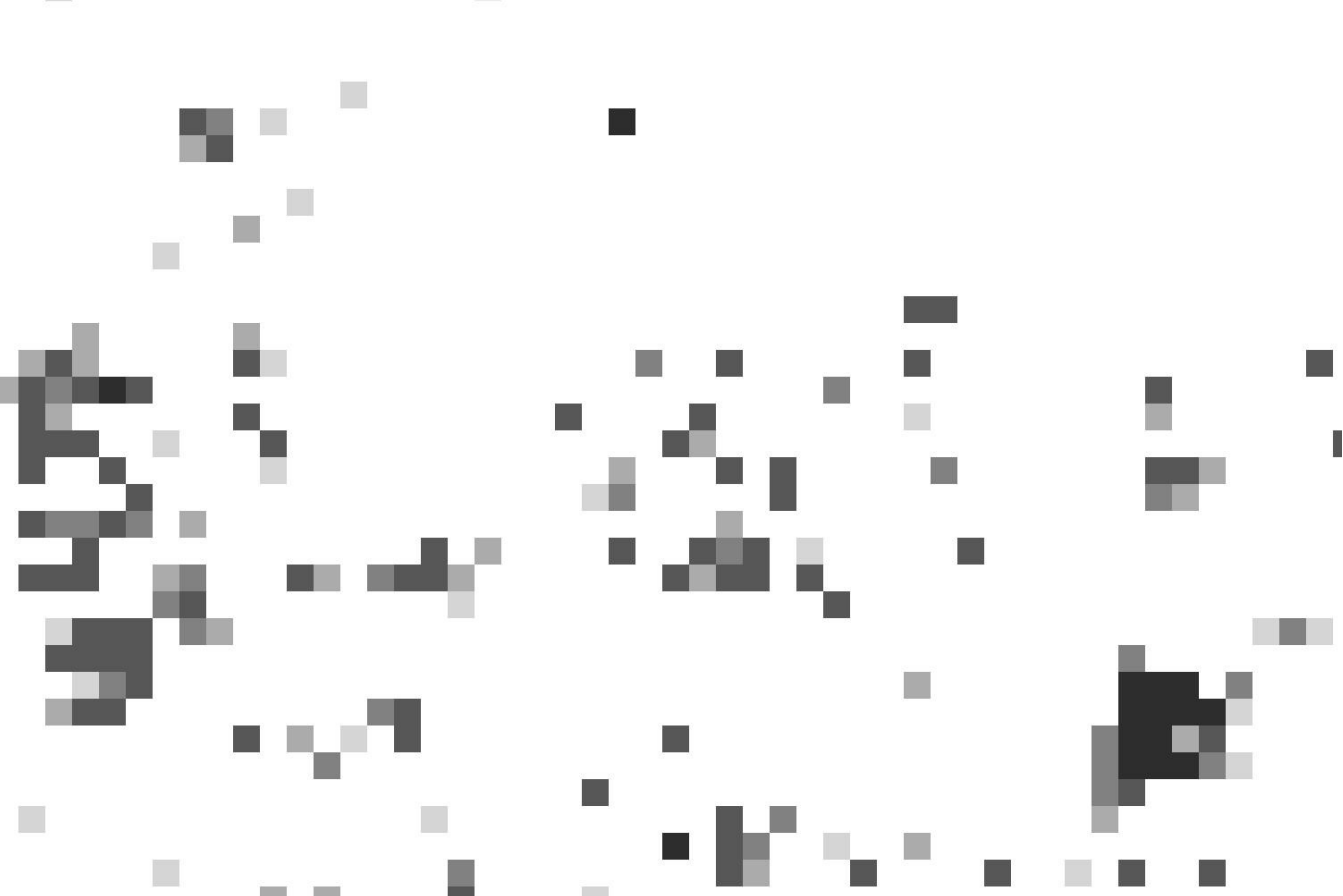}
\end{center}
\caption{Figure, in the right-hand side, shows the amount of processing (number of stages used in the cascade) in order to reject or accept different patches in the test image (shown in the left-hand side); darker colors correspond to more intense processing.}\label{fig4}
\end{figure*}

\noindent  Fig.~(\ref{fig1}, right) and Table.~(\ref{tab2}) show respectively the evolution of the objective function (\ref{eq1}) w.r.t different iterations of optimization and different evaluation measures of the underlying $g$-networks (obtained at convergence) w.r.t different stages of the cascade. We observe from these results that as we go deep in the cascade, the characteristics of the $g$-networks resemble more and more the original $f$-network; the efficiency decreases and the discrimination power (EER) remains stable or improves.\\
Fig.~\ref{fig3} shows examples of change detection results obtained through different stages of the cascade and Fig.~\ref{fig4} shows the amount of processing in order to classify different patches as changes or no-changes.  From these results, it is clear that almost all the areas are rejected at the early stages of the cascade and only few areas (changes and change-like structures) require more intense processing. In practice, our coarse-to-fine cascade  is almost $10\times$ faster than the original $f$-network while its overall EER (shown in Table.~\ref{tab3}) remains relatively stable.  
 
\begin{table}
\begin{center}
 \resizebox{0.7\columnwidth}{!}{
\begin{tabular}{c||cc||cc||cc}
Stage &  \%cons & \%rFA& \%DR & \%FA & \%ERR &  time(ms) \\
  \hline
  \hline
1 & 99.16  &  41.04   &97.55  &  43.56    & 23.00  &745\\
2 & 98.86  &  37.37   &96.88 &   40.09      &  21.61  &897 \\
3 & 98.77 &    35.27  &96.70  &  38.07 &  20.68  &1180 \\ 
4 & 98.80 &    39.78  &96.76 &   42.53  & 22.89   & 2721 \\
5 & 98.70 &    28.24  &96.56 &   31.46  &  17.45  &4639 \\ 
6 & -   &  -   & {\bf 97.14}   & {\bf 04.44}   & {\bf 03.65}  & 14230
\end{tabular}}
\vspace{0.1cm}
\caption{This table shows different evaluation measures of the $g$-networks w.r.t stages of the cascade as well as the average processing time. All these percentages are evaluated on the test set $\cal U$ while processing time is the average time to process a given pair of very large reference and test images (of $9850 \times 10400$ pixels);   each reference and test image  includes $113488$ patches. Note that cons and rFA are  not given for stage 6 as the $g$-network of this stage is exactly the $f$-network so these measures are obviously equal to 100\% and 0\% respectively. In order to study the behavior of different stages independently, each classifier is evaluated using all the data in ${\cal U}$, so the resulting FAs are not necessarily decreasing. In these experiments, the significant increase of FAs (from stage 6 to the other stages) is mainly due to the implementation of the conservation hypothesis that maintains a high detection rate to the detriment of an increase of false alarms.}\label{tab2}
\end{center}
\end{table}

\begin{table}
\begin{center}
 \resizebox{0.7\columnwidth}{!}{
\begin{tabular}{c||ccc|c}
 & \%DR & \%FA & \%EER  & time (ms)   \\
  \hline
  \hline
$f$-network+classifier  & {97.14}  &  04.44  & {\bf 03.65}  &  14230 \\
cascade &  92.16 & {03.80}  & {\bf 05.82}   &     1627 ($\sim10\times $ faster)                                                   
\end{tabular}}
\vspace{0.1cm}
\caption{This table shows the overall performances of the $f$-network and the cascade of networks. Again, these percentages are evaluated on the test set $\cal U$ while processing time is the average time to process  a given pair of very large reference and test images (of $9850 \times 10400$ pixels);  each reference and test image includes  $113488$ patches.}\label{tab3}
\end{center}
\end{table}

\section{Conclusion}\label{section5}

We introduced in this paper a novel approach for efficient deep kernel network evaluation.  The design principle of our method is coarse-to-fine; it is based on a cascade of kernel networks and classifiers with increasing complexity and discrimination power. Networks in the early stages of the cascade are cheap and are used to reject many patterns efficiently while those belonging to the deep stages of the cascade are more expensive and more discriminating.  The parameters of these networks are obtained by solving  several cross-entropy minimization problems that reduce the difference between the original and the reduced cost kernel networks.\\  
 Even though tested on the particular (challenging) problem of change detection, this method is generic and could be extended to other imbalanced classification tasks such as  object and rare event detection in  images and videos where the untargeted  classes are dominant. Other possible extensions of this work include transfer learning; indeed one may reduce the complexity of existing very deep networks (using our optimization framework) prior to achieve fine-tuning as this may reduce the computational cost of learning very significantly.

{
\bibliographystyle{IEEEtran}
\bibliography{egbib}
}
 
\end{document}